\documentclass[10pt,twocolumn,letterpaper]{article}

\usepackage[pagenumbers]{cvpr} 

\usepackage{tabularx} 

\usepackage{booktabs}
\usepackage{subcaption}
\usepackage{adjustbox}
\usepackage{caption} 
\usepackage{multirow}

\definecolor{cvprblue}{rgb}{0.21,0.49,0.74}
\usepackage[pagebackref,breaklinks,colorlinks,allcolors=cvprblue]{hyperref}

\newcommand{\bc}{\mathbf{c}}
\newcommand{\bj}{\mathbf{j}}
\newcommand{\bp}{\mathbf{p}}
\newcommand{\bt}{\mathbf{t}}

\newcommand{\bv}{\mathbf{v}}
\newcommand{\bomega}{\boldsymbol{\omega}}
\newcommand{\cL}{\mathcal{L}}
\newcommand{\tsim}{\text{sim}}
\newcommand{\timg}{\text{img}}

\def\confName{CVPR}
\def\confYear{2026}

\title{Grasp-and-Lift: Executable 3D Hand-Object Interaction Reconstruction via Physics-in-the-Loop Optimization
}

\author{
    Byeonggyeol Choi$^{1}$\thanks{Equal contribution} \quad Woojin Oh$^{2\ast}$ \quad Jongwoo Lim$^{1,2}$\thanks{Corresponding author} \\
    $^{1}$ME \quad $^{2}$IPAI \\
    Seoul National University \\
    {\tt\small \{skymap33, woojin.oh, jongwoo.lim\}@snu.ac.kr}
}

\begin{document}
\maketitle
\begin{abstract}

Dexterous hand manipulation increasingly relies on large-scale motion datasets with precise hand-object trajectory data. However, existing resources such as DexYCB and HO3D are primarily optimized for visual alignment but often yield physically implausible interactions when replayed in physics simulators, including penetration, missed contact, and unstable grasps. 

We propose a simulation-in-the-loop refinement framework that converts these visually aligned trajectories into physically executable ones. Our core contribution is to formulate this as a tractable black-box optimization problem. We parameterize the hand's motion using a low-dimensional, spline-based representation built on sparse temporal keyframes. This allows us to use a powerful gradient-free optimizer, CMA-ES, to treat the high-fidelity physics engine as a black-box objective function. Our method finds motions that simultaneously maximize physical success (e.g., stable grasp and lift) while minimizing deviation from the original human demonstration. 

Compared to MANIPTRANS-recent transfer pipelines, our approach achieves lower hand and object pose errors during replay and more accurately recovers hand-object physical interactions. Our approach provides a general and scalable method for converting visual demonstrations into physically valid trajectories, enabling the generation of high-fidelity data crucial for robust policy learning.

\end{abstract}    
\section{Introduction}
\label{sec:intro}

\begin{figure}[t!]
  \centering
  \includegraphics[width=\linewidth]{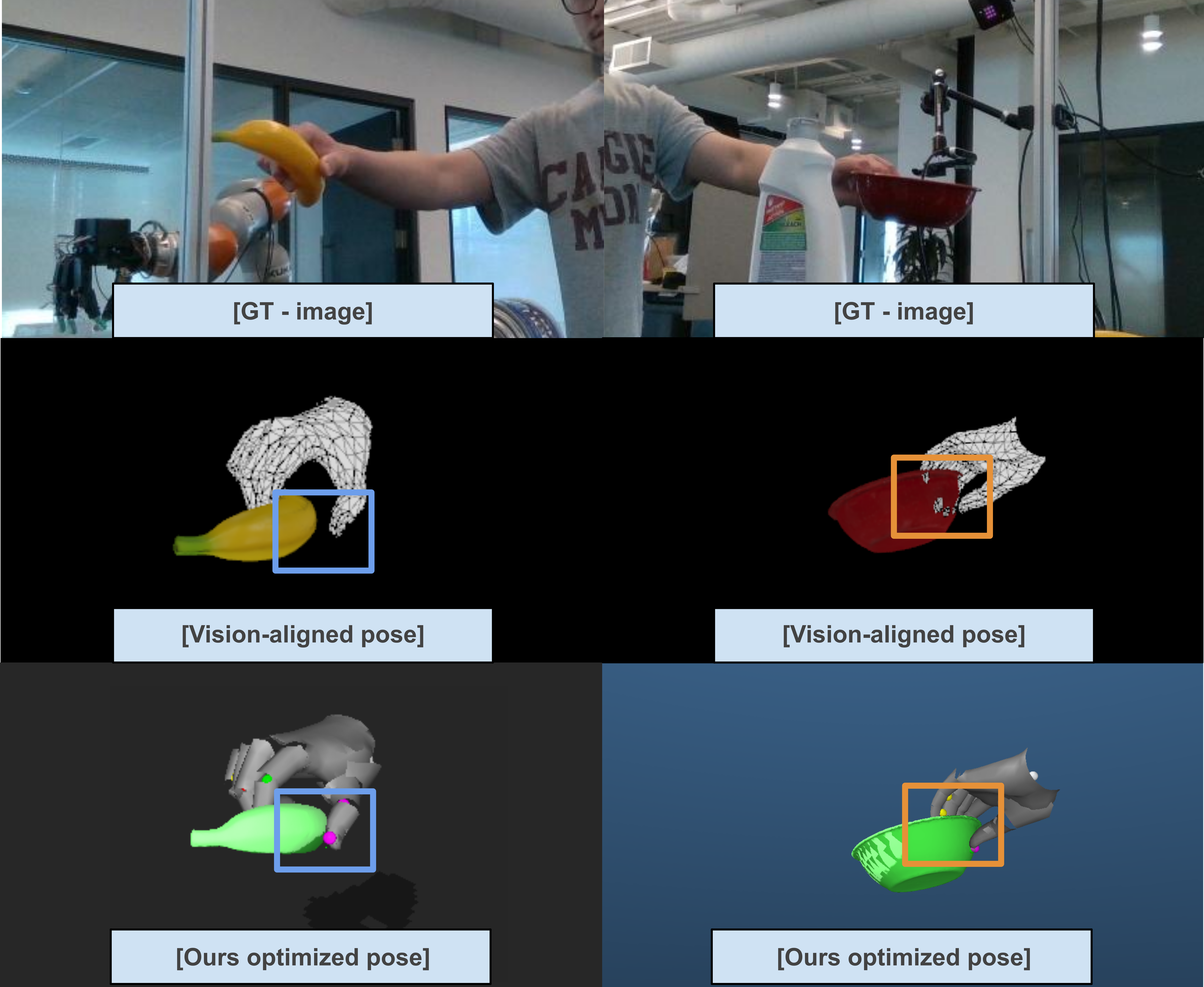}
  \caption{Examples of our physics-based refinement. The original vision-aligned dataset (DexYCB)~\cite{Chao2021DexYCB} often contains physical impossibilities, such as significant gaps (blue boxes) or hand-object interpenetration (orange boxes). Our optimization framework refines these poses into physically plausible interactions, promoting valid and stable hand-object contact.}
  \label{fig:objective}
  \vspace{-1em}
\end{figure}

Dexterous robotic manipulation is a cornerstone of next-generation robotics, but training such systems requires large-scale, high-fidelity data on human-object interactions. Publicly available datasets, such as DexYCB~\cite{Chao2021DexYCB}, and HO-3D~\cite{Hampali2020HO3D}, have been essential in providing high-quality, vision-aligned hand and object pose annotations across diverse tasks. However, these datasets are optimized for visual consistency rather than physical realism. When their trajectories are replayed in a physics simulator, the underlying ``vision-physics mismatch" becomes apparent: interactions often exhibit significant non-physical behaviors such as interpenetration, missed contacts, or unstable grasps (See Fig.~\ref{fig:objective}). This mismatch causes the actual motion of the object to diverge drastically from the original demonstration when the action is executed, which limits the data's utility for training contact-aware manipulation policies.

Force-torque sensors or sensorized gloves are widely used devices to capture this missing physical ground truth. However, this approach is often impractical for large-scale data collection. It faces several issues: high sensor costs, intrusive mounting (which can alter object geometry and dynamics), visual occlusions from markers and cabling, and persistent challenges with calibration drift and sensor wear. Therefore, there is a critical need for a method that can recover and validate physical interactions directly from the vast, existing, vision-only datasets.
To bridge this gap, we propose a ``real-to-sim" refinement framework. Our goal is to transform these \emph{visually plausible but physically flawed} demonstrations into \emph{physically executable} trajectories. Given a vision-only hand-object sequence, our framework optimizes it to find a new, dynamically consistent motion that is successful in simulation while remaining faithful to the original human demonstration.
The resulting trajectories are not just visually plausible but also dynamically validated. A significant benefit of our physics-based approach is that we recover explicit 6-DoF contact wrenches (forces and torques) and contact positions as a free by-product, all without requiring any sensors. In experiments, our framework achieves much lower hand and object trajectory errors and faster optimization than recent pipelines like MANIPTRANS~\cite{Li2025ManipTrans}. The refined rollouts serve as a high-quality benchmark for contact-aware policy learning, narrowing the gap between human demonstrations, simulations, and real-world deployments.
\noindent
Our work makes three main contributions:
\begin{itemize}
\item \textbf{Novel Physics-in-the-Loop Refinement Framework} \\
We propose a ``real-to-sim" framework that formulates trajectory refinement as a control-based optimization problem. Instead of optimizing object poses, our method optimizes only the hand's joint parameters, allowing the object's motion to emerge directly from the interaction of the physical simulation. This approach inherently guarantees dynamic consistency.

\item \textbf{Efficient and Accurate Optimization Strategy} \\
To solve this complex, high-dimensional control problem, we introduce a spline-based temporal keyframe parameterization. This technique dramatically reduces the search space and enforces smoothness, enabling our gradient-free optimizer (CMA-ES~\cite{hansen2016cma}) to converge significantly faster and achieve lower hand and object trajectory errors compared to the state-of-the-art method MANIPTRANS~\cite{Li2025ManipTrans}.

\item \textbf{Force and Contact Estimation from Vision-only Data} \\
Our method is entirely vision-based, and requires no motion capture or force/torque sensors. As a direct byproduct of its physics-based optimization, it generates explicit 6-DoF contact wrenches (forces and torques) and time-aligned contact positions, providing a valuable source of rich, physical supervision for contact-aware manipulation policy learning.



\end{itemize}

\section{Related Works}
\label{sec:relatedworks}

\subsection{Human-to-Robot Transfer}

A line of work transfers human hand skills to robot embodiments via retargeting, imitation, or policy learning. MANIPTRANS~\cite{Li2025ManipTrans} proposes a two-stage pipeline (trajectory imitation + residual fine-tuning) to transfer bimanual human demonstrations to dexterous robot hands in simulation, emphasizing accurate tracking and data efficiency. DexMV~\cite{Shen2022DexMV} converts human videos into robot demonstrations for dexterous tasks through vision-to-simulation translation and imitation learning. VideoDex~\cite{Shaw2023VideoDex} mines large-scale internet human-hand videos to provide priors for dexterous manipulation, learning general-purpose skills from in-the-wild footage. Recent works extend human-to-robot skill transfer with improved plausibility constraints directly from human motion, focusing on generating robot actions that are dynamically feasible~\cite{HumanRobot2025Plausible}. Classical retargeting pipelines also map human RGB-D hand motion to robot hand kinematics, highlighting practical recipes for demonstration acquisition~\cite{TUMRetargeting2020}.



\subsection{Estimation of Contact and Forces}

To bridge from vision to physically meaningful interaction, ContactPose~\cite{ContactPose2020} and ContactDB~\cite{brahmbhatt2019contactdb} provide contact maps paired with hand/object pose, enabling learning of contact predictors and analysis of grasp contact statistics. Optimization-based ContactOpt~\cite{Grady2021ContactOpt} refines hand poses to realize predicted contact distributions with differentiable contact models. Beyond static contact, several works infer forces from vision: early efforts estimate manipulation forces from monocular video of human-object interactions, while more recent approaches tackle vision-based interaction force estimation for robotic grasping~\cite{Li2019Forces}. Broader physics-based reconstruction uses trajectory optimization with contact constraints to recover physically valid motions from video, underscoring the importance of dynamic consistency in reconstructions~\cite{PhysicsRecon2020}. Complementary pipelines reconstruct object pose/shape with hand interactions from RGB sequences, informing contact-aware perception and downstream control~\cite{Hampali2023InHandScan}. Closest in spirit, G\"artner et al.~\cite{Gartner2022PhysicsHuman} perform trajectory optimization with a full physics engine to reconstruct physically plausible 3D human motion from monocular video, demonstrating generalization beyond controlled labs and handling rich contact events---an approach conceptually aligned with our physics-constrained refinement.

\subsection{Physical-Plausibility Optimization}

Beyond vision-only retargeting, a large body of work directly optimizes grasp poses (and short trajectories) under contact physics. 
Grasp’D~\cite{Turpin2022GraspD} formulates differentiable contact-rich grasp synthesis and shows that reasoning about contact geometry improves grasp quality and stability under simulation. 
Fast-Grasp’D~\cite{Turpin2023FastGraspD} pushes this to real-time generation with GPU-accelerated pipelines. 
ArtiGrasp~\cite{Zhang2024ArtiGrasp} synthesizes physically plausible bi-manual grasps while coordinating articulation of hands and objects, emphasizing realism under contact constraints. 
Earlier graphics/robotics works established contact-aware trajectory optimization for dexterous manipulation: 
Liu et al.~\cite{Liu2009DexterousManipulationFromGrasp} optimize post-grasp dexterous manipulation from a given grasp; 
and Contact-Invariant Optimization (CIO) by Mordatch et al.~\cite{Mordatch2012CIO} discovers complex contact sequences by optimizing trajectories with contact costs that turn on/off smoothly. 
Together, these works optimize for physically valid grasps and short interaction snippets; 
our approach complements them by refining full H--O trajectories from vision and recovering explicit contact signals (positions/forces) without sensors, while preserving visual alignment.

\begin{figure*}[t!]
  \centering
  \includegraphics[width=\textwidth]{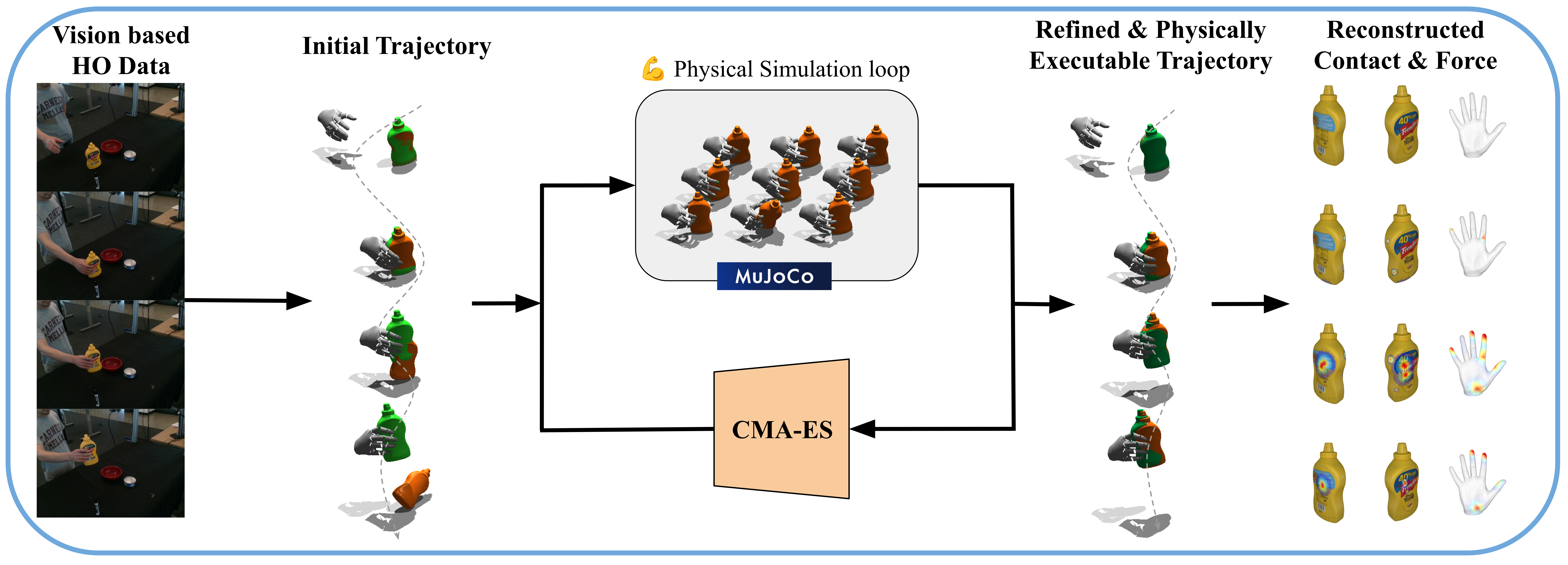}
  \caption{\textbf{Overall pipeline.} Given vision-aligned object 6D pose and hand pose estimated from multi-view images (e.g., DexYCB), we replay these trajectories in a physics simulator to expose interaction cues (contacts, normals, forces, penetrations), which define physics-informed losses. A black-box optimizer (CMA-ES) then updates a low-dimensional control vector—top-10 MANO PCA coefficients plus wrist rotation in $\mathrm{SO}(3)$—to refine \emph{the hand trajectory only}, while the object's motion is updated as a result of the simulated physical interaction. The result is a replayable, physically plausible hand–object interaction that remains consistent with image evidence and provides explicit contact positions and forces for downstream use.}
  \label{fig:overview}
\end{figure*}

\section{Method}
\label{sec:method}

We present a simulation-in-the-loop framework to refine visually aligned hand-object demonstrations into physically executable trajectories. Figure~\ref{fig:overview} illustrates an overview of our method. Our approach reformulates this task as a control-based optimization problem. Instead of directly optimizing hand and object poses, our method optimizes only the hand's joint parameters, which serve as control input to the simulator. The object's motion is not prescribed; it emerges as a direct, dynamic consequence of the simulated physical contacts and forces.

To make this high-dimensional optimization problem tractable and enforce temporal smoothness, we do not optimize every frame. Instead, we parameterize the hand's control trajectory using a cubic spline interpolation over a sparse set of temporal keyframes. This spline-based representation dramatically reduces the search space. We then treat the non-differentiable physics engine as a black-box objective function. We use the Covariance Matrix Adaptation Evolution Strategy (CMA-ES~\cite{hansen2016cma}), a powerful zeroth-order optimizer, to find the optimal keyframe parameters that minimize a sequence-level loss. This objective ensures that the resulting motion is physically consistent and remains faithful to the original demonstration. We instantiate our framework in MuJoCo~\cite{2012Mujoco}, leveraging its fast and stable contact dynamics. 

\subsection{Hand-Object Trajectory Parameterization}
\label{sec:handobjparam}

\paragraph{Hand Pose Parameterization}
Directly optimizing the hand pose using per-joint angles presents a high-dimensional search space (e.g., 21 DoF). This approach is not only sample-inefficient for a black-box optimizer, but also prone to anatomically implausible poses that violate joint limits or cause self-collisions.

To address these issues, we leverage the low-dimensional latent space of the MANO hand model. We parameterize the hand’s articulation using its 45-component principal component analysis (PCA) basis, but for optimization we use a reduced set consisting of the top 10 PCA coefficients and the wrist rotation ($\mathrm{SO}(3)$). \textit{Concretely, the hand’s control vector is} $\mathbf{u}_t^{\text{hand}} = [\,\boldsymbol{\theta}_t,\; R_t^{\text{wrist}}\,]$, \textit{which combines the MANO PCA coefficients and the wrist rotation.} This low-dimensional, data-driven parameterization inherently preserves anatomical plausibility, substantially reduces the optimizer’s workload, and stabilizes and accelerates convergence, enabling efficient CMA-ES within MuJoCo.

Moreover, human hands are substantially more flexible than robot hands: soft-tissue compliance and skin sliding can induce meaningful shape changes even away from explicit joints (e.g., palm bulging and local skin motion when squeezing an object). Optimizing in MANO’s PCA space captures these correlated, non-rigid variations in a compact, low-rank form without introducing extra joint parameters. In simulation, we decode PCA coefficients back to an articulated MANO rig (``Arti-MANO'') with $\sim$45 articulated DoFs and use this as the MuJoCo hand asset, following the practice introduced in \cite{Li2025ManipTrans}. This practice itself builds on the original MANO formulation \cite{MANO2017} and subsequent physics-based articulated MANO pipelines such as D-Grasp \cite{Christen2022DGrasp}. In our implementation, we start from such an articulated MANO hand and, when beneficial, augment it with a small set of auxiliary DoFs—\emph{e.g.}, limited axial-rotation (“roll”) near finger bases or comparable proxies for soft-tissue accommodation—so as to better capture subtle in-hand adjustments during contact, yielding natural, human-like motions that remain faithful to vision evidence while being well-suited for accurate, physics-based contact reasoning.

\paragraph{Reference Object Trajectory}
We assume that an accurate mesh model of the object is known. The algorithm is given the object's 6D pose trajectory estimated from the visual data, but this trajectory is not treated as a fixed kinematic target. Instead, it serves as the target reference trajectory for our optimization.

During the simulation-in-the-loop optimization, the simulator dynamically computes the object's motion based on the contact forces applied by the optimized hand pose. The goal of our framework is to find a hand control sequence that causes the object's simulated motion to match the target visual motion. Our objective functions (detailed in Section~\ref{sec:objective}) enforce this alignment by penalizing deviations between the pose and velocity of the simulated object and the original visual track. This formulation anchors the optimization, ensuring that the refined, physically executable interaction remains faithful to the original human demonstration.

\paragraph{Temporal Keyframe Optimization.}
Optimizing the hand's control parameters at every single frame of a long sequence is computationally prohibitive and prone to overfitting or high-frequency jitter. We therefore adopt a keyframe-based temporal parameterization.

Specifically, we select a keyframe every five frames. The control parameters for all intermediate frames are then generated using cubic spline interpolation. This spline coupling is a critical component of our method:
\begin{itemize}
\item It enforces temporal smoothness by ensuring continuous, low-jerk motion for both the wrist and finger articulations.
\item It also dramatically reduces the number of free variables that  the CMA-ES must optimize, making the high-dimensional control problem tractable.
\end{itemize}
To maintain continuity across the entire sequence, we employ a sliding-window strategy, optimizing the trajectory in overlapping segments. This formulation yields smooth, consistent, and physically plausible motion while keeping the optimization efficient. 


\subsection{Objective Functions}
\label{sec:objective}

To find the visually plausible and physically executable hand trajectory, we minimize a comprehensive sequence-level objective $\cL(\mathbf{u}_t^{\text{hand}})$. This objective is a weighted sum of several cost functions designed to enforce two key properties: physical realism (i.e., the simulated motion is physically valid and matches the object's dynamics) and visual fidelity (i.e., the refined motion remains faithful to the original human demonstration). The total loss is aggregated over the time sequence $t$,
\begin{equation*}
\begin{aligned}
\cL(\mathbf{u}_t^{\text{hand}})=\sum_t^{}\big[&
\lambda_\text{op}\cL_\text{obj-pose}^{(t)}+
\lambda_\text{ov}\cL_\text{obj-vel}^{(t)}+
\lambda_\text{hj}\cL_\text{hand-joint}^{(t)}\\
&+\lambda_\text{ct}\cL_\text{contact}^{(t)}+
\lambda_\text{hc}\cL_\text{hand-cons}^{(t)}+
\lambda_\text{hv}\cL_\text{hand-vel}^{(t)}\big],
\end{aligned}
\end{equation*}
where the components are defined as follows.

\paragraph{Object Pose Loss $\cL_\text{obj-pose}$:}
This loss is our primary objective, ensuring the simulated object's motion from the interaction with hand, matches the target visual trajectory. We penalize the difference between the simulated pose $(R_t^\tsim, \bt_t^\tsim)$ and the visual reference pose $(R_t^\timg, \bt_t^\timg)$ using geodesic distance for rotation and $L_2$ norm for translation.
\begin{equation*}
\begin{aligned}
&d_\mathrm{geo}(R_1, R_2) = \arccos\!\left(\tfrac{\mathrm{tr}(R_1 R_2^\top)-1}{2}\right), \\[6pt]
&\cL_\text{obj-pose}^{(t)} =
\eta\, d_\mathrm{geo}(R_t^\tsim, R_t^\timg)
\,+\,(1-\eta)\,\|\,\bt_t^\tsim-\bt_t^\timg\,\|_2,
\end{aligned}
\end{equation*}
where $(R^\tsim_t,\bt^\tsim_t) \in SE(3)$ denotes the object pose in simulation and $(R^\timg_t,\bt^\timg_t)$ is the image-aligned object pose from vision data.

\paragraph{Object Velocity Loss $\cL_\text{obj-vel}$:}
To better enforce dynamic consistency, we also penalize the difference between the simulated $(\bv_t^\tsim, \bomega_t^\tsim)$ and target $(\bv_t^\timg, \bomega_t^\timg)$ linear and angular velocities. This loss complements the pose term by ensuring that the object moves in a physically plausible way from frame to frame,
\begin{equation*}
\cL_\text{obj-vel}^{(t)}=
\lambda_{ov}\|\mathbf v_t^\tsim-\mathbf v_t^\timg\|_2^2+
\lambda_{o\omega}\|\boldsymbol\omega_t^\tsim-\boldsymbol\omega_t^\timg\|_2^2.
\end{equation*}

\paragraph{Hand 3D Joint Position Loss $\cL_\text{hand-joint}$:}
This loss acts as a visual prior, guiding the hand's simulated pose $(\bj_{t,i}^\tsim)$ to stay near the original visual demonstration $(\bj_{t,i}^\timg)$, where $\bj_{t,i}$ is the 3D position of the hand joint $i$ at time $t$. We up-weight the fingertips $(w_{tip} > 1)$, as they are the primary source of interaction,
\begin{equation*}
\cL_\text{hand-joint}^{(t)} =
\frac{1}{W}\sum_{i=1}^{20}w_i\,\|\,\bj^\tsim_{t,i}-\bj^\timg_{t,i}\,\|_2,
\end{equation*}
\begin{equation*}
\text{where~}
w_i =
\begin{cases}
w_\mathrm{tip} & \text{for fingertips}\\
1 & \text{otherwise}
\end{cases}
\text{~and~}
W = \sum_{i=1}^{20} w_i.
\end{equation*}

\paragraph{Hand-Object Contact Loss $\cL_\text{contact}$:}
To further guide the executable physical interaction, we pre-compute likely contact pairs from the visual data. For each frame, we identify fingertips $(k)$ that are close to the object surface and store their corresponding object-local contact points $(\bc_{t,k}^\text{img,obj})$. This loss then penalizes the distance between the simulated fingertip position $(\bp_{t,k}^\text{sim,hand})$ and this target contact point, which is transformed by the simulated object's current pose,
{\small \begin{equation*}
\cL_\text{contact}^{(t)}=
\frac{1}{\max\!\big(1,|\mathcal{K}_t|\big)}
\sum_{k\in\mathcal{K}_t}
\Big\|
\mathbf p^\text{sim,hand}_{t,k}
-\big(R_t^\tsim\mathbf c^\text{img,obj}_{t,k}+\bt_t^\tsim\big)
\Big\|_2,
\end{equation*}}
where $\mathcal{K}_t$ is the set of object-fingertips matched at frame $t$.

\paragraph{Hand Regularization Losses $\cL_\text{hand-cons}$ \& $\cL_\text{hand-vel}$:} 
Finally, to ensure smooth and stable motion, we apply two regularization terms to the hand trajectory. The consistency loss ($\mathcal{L}_{\text{hand-cons}}$) penalizes frame-to-frame 
changes in the the hand control vector $\mathbf{u}_t^{\text{hand}}$, suppressing 
jitter introduced by the sampling process during optimization.
This term suppresses jitter from the sampling process during optimization. The velocity loss ($\cL_\text{hand-vel}$) penalizes excessive linear and angular velocities of the hand's bodies $\mathcal{B}$, further encouraging smooth motion:
\begin{equation*}
\mathcal{L}_{\text{hand-cons}}^{(t)} =
\bigl\|\boldsymbol{\theta}_t - \boldsymbol{\theta}_{t-1}\bigr\|_2^2
\;+\;
d_{\mathrm{geo}}\!\bigl(R_t^{\text{wrist}},\, R_{t-1}^{\text{wrist}}\bigr),\text{and}
\end{equation*}
%
\begin{equation*}
\cL_\text{hand-vel}^{(t)}=
\lambda_{hv} \sum_{b\in\mathcal B} \big\|\mathbf v_{t,b}\big\|_2^2
\;+\;
\lambda_{h\omega} \sum_{b\in\mathcal B} \big\|\boldsymbol\omega_{t,b}\big\|_2^2.
\end{equation*}


\subsection{Optimizer and Simulator Setup}

\subsubsection{Physics Simulator}
Contact-rich manipulation induces stiff, non-smooth dynamics, such as impacts and stick-slip transitions. While differentiable simulators expose gradients by smoothing or relaxing these events (e.g., using compliant contact models), the resulting gradients are often ill-conditioned near mode switches and may not capture sharp impulses.

Our goal is not to backpropagate through dynamics, but to recover physically plausible and replayable trajectories with high-fidelity contact signals. We therefore select MuJoCo as our physics engine. MuJoCo provides fast, stable rigid-body contact resolution with tunable compliance and, crucially, allows for explicit per-contact queries (positions on both bodies, normals, and 6-DoF wrenches), which are essential for generating our ground-truth contact data. 

\subsubsection{Zeroth-order Optimizer Selection}
Our objective function (Section~\ref{sec:objective}) is piecewise smooth and non-differentiable in the hand parameters due to the discrete nature of contact events (contacts appearing/disappearing) and friction mode switches.

We therefore use the Covariance Matrix Adaptation Evolution Strategy (CMA-ES), a state-of-the-art derivative-free optimizer that is highly robust to such discontinuities and simulator noise. CMA-ES is naturally compatible with our simulation-in-the-loop evaluation, as it requires no dynamics gradients. It is particularly well-suited to our method, as it efficiently searches the low- to medium-dimensional control space defined by our spline-based MANO-PCA parameterization (Section~\ref{sec:handobjparam}).

\subsubsection{Windowed Optimization Schedule}
To optimize the full trajectory, we employ a sliding-window strategy over the keyframes defined in Section~\ref{sec:handobjparam}. A brute-force optimization of all keyframes at once would be inefficient and unstable. Instead, we form an optimization window using four consecutive keyframes (e.g., $k_1, k_2, k_3, k_4$). 

To create a well-posed problem that respects the temporal context, we fix the ``past" keyframes ($k_1, k_2$) and the ``future" keyframe ($k_4$). CMA-ES is then used to optimize the parameters of only the third keyframe ($k_3$). 
This strategy is key to our method's stability. Because the cubic spline is coupled across all four points, the optimizer must find a pose for $k_3$ that smoothly interpolates from the fixed past while also anticipating the fixed future. The per-window loss aggregates terms from all frames in this window, enforcing this anticipatory behavior (e.g., shaping the grasp posture before contact to ensure a stable lift later). After converging on an optimal $k_3$, we slide the window forward by one keyframe and repeat the process, propagating the spline-smoothed solution across the entire sequence. 

\section{Experiments}
\label{sec:experiments}

In this section, we first describe the experimental setup, including the datasets, evaluation metrics, and implementation details (Sec.~\ref{subsec:exp_setup}). We then present a comparative analysis against state-of-the-art (SOTA) methods (Sec.~\ref{subsec:results}), qualitative analysis of physics reconstruction (Sec.~\ref{subsec:qualitative}), and ablation studies (Sec.~\ref{subsec:ablation}).

\subsection{Experimental Setup}
\label{subsec:exp_setup}

\paragraph{Dataset and Baseline}

We use the DexYCB~\cite{Chao2021DexYCB} dataset for all experiments. Since the complete dataset is extensive, we compiled a challenging evaluation set of 120 sequences. This set is designed to test generalization by sampling three different subjects, 20 YCB objects, and a wide variety of grasp types. Each sequence at 30 fps is 60–70 frames long. We exclude the first 10 frames to ensure a valid interaction.

Our primary baseline is MANIPTRANS, a state-of-the-art, policy-based transfer method. We used their official, public implementation and developed a custom DexYCB data loader to enable fair comparisons using the same setup.

\paragraph{Evaluation Metrics}

For a fair comparison, we adopt the primary metrics from MANIPTRANS and also report \emph{total optimization time} for both methods. 

\begin{itemize}
\item \textbf{Object Pose Error:} We measure the average rotation error ($E_r$, in degrees) and translation error ($E_t$, in cm) between the simulated object pose and the image-aligned reference pose:\vspace{-.5em}
{\small\[
E_r = \frac{1}{T} \sum_{t=1}^{T} d_\mathrm{geo}(R_t^\tsim, R_t^\timg),\;\;
E_t = \frac{1}{T} \sum_{t=1}^{T} \left\|\, \bt_t^\tsim - \bt_t^\timg \,\right\|_{2}.
\]}
\item \textbf{Hand Pose Error:} We report the Mean Per-Joint Position Error ($E_j$, in cm) and the Mean Per-Fingertip Position Error ($E_{ft}$, in cm):\vspace{-.5em}
{\small\[
E_j = \frac{1}{T \cdot F} \sum_{t=1}^{T} \sum_{f=1}^{F} \left\|\, \bj_{t,f}^\tsim - \bj_{t,f}^\timg \,\right\|_{2},
\]\vspace{-1em}
\[
E_{ft} = \frac{1}{T \cdot M} \sum_{t=1}^{T} \sum_{m=1}^{M} \left\|\, \bj_{t,m}^\tsim - \bj_{t,m}^\timg \,\right\|_{2}.
\]}
\item \textbf{Success Rate (SR):} We use the MANIPTRANS failure thresholds (e.g., $E_r > 30^\circ$, $E_t > 3$ cm). As our optimization framework is deterministic, a sequence's SR is binary (100\% if all frames pass, 0\% if any frame fails).
\end{itemize}

\paragraph{Implementation Details}
We instantiate our framework in MuJoCo and use the official object mass specifications from DexYCB to ensure physical accuracy. For the MANIPTRANS baseline, we set its IsaacGym simulation~\cite{makoviychuk2021isaac} timestep to $1/30$s to match the DexYCB capture rate. All experiments were conducted on a workstation with an NVIDIA RTX A6000 GPU and an Intel i7-14700K CPU.



\begin{table*}[t!]
\centering
\small
\begin{tabular}{l|cccccc|cccccc}
\hline\hline
\multirow{2}{*}{Subjects} &
\multicolumn{6}{c|}{\textbf{ManipTrans}} &
\multicolumn{6}{c}{\textbf{Ours}} \\
\cline{2-13}
& Time (s) & $E_t$ & $E_r$ & $E_j$ & $E_{ft}$ & SR (\%)
& Time (s) & $E_t$ & $E_r$ & $E_j$ & $E_{ft}$ & SR (\%) \\
\hline
Subject-1
& 1546.86 & 0.88 & 4.01 & 0.99 & 1.20 & 61.76
& \textbf{533.24} & \textbf{0.54 }& \textbf{2.26} & \textbf{0.98} & \textbf{1.18} & \textbf{85.0} \\
Subject-2
& 2184.10 & 0.97 & 12.07 & 2.55 & 2.17 & 55.23
& \textbf{583.57} & \textbf{0.72} & \textbf{2.67} & \textbf{1.09} & \textbf{1.42} & \textbf{77.5} \\
Subject-3
& 2229.77 & 0.95 & 9.43 & 2.65 & \textbf{2.26} & 20.79
& \textbf{554.04} & \textbf{0.75} & \textbf{3.98} & \textbf{2.58} & 2.93 & \textbf{52.5} \\
\hline
Total(Avg.)
& 1986.91 & 0.93 & 8.50 & 2.06 & 1.87 & 45.93
& \textbf{556.95} & \textbf{0.67} & \textbf{2.97} & \textbf{1.55} & \textbf{1.84} & \textbf{71.67} \\
\hline\hline
\end{tabular}
\caption{\textbf{Per-subject quantitative comparison on DexYCB.} Each subject
corresponds to a distinct human demonstrator, so all sequences in a row share
the same operator. Across all subjects, our method is about $3{\sim}4\times$
faster to optimize than ManipTrans and attains lower pose errors on both hand and object with a
substantially higher success rate. (Note: $E_t, E_j, E_{ft}$ are in cm and
$E_r$ is in degrees.)}
\label{tab:quant_per_subject}
\end{table*}

\begin{table*}[t!]
\centering
\small
\begin{tabular}{l|cccccc|cccccc}
\hline\hline
\multirow{2}{*}{Object} &
\multicolumn{6}{c|}{\textbf{ManipTrans}} &
\multicolumn{6}{c}{\textbf{Ours}} \\
\cline{2-13}
& Time (s) & $E_t$ & $E_r$ & $E_j$ & $E_{ft}$ & SR (\%)
& Time (s) & $E_t$ & $E_r$ & $E_j$ & $E_{ft}$ & SR (\%) \\
\hline
002\_master\_chef\_can  &  1937.92  &  0.58  &  4.21  &  1.81  &  1.92  &  49.74  &  \textbf{641.12} &  \textbf{0.42}  &  \textbf{1.47}  &  \textbf{0.85}  &  \textbf{1.06}  &  \textbf{83.33} \\
003\_cracker\_box  &  2018.04  &  0.50  &  4.09  &  1.02  &  1.08  &  33.33  &  \textbf{515.42} &  \textbf{0.41}  &  \textbf{1.29}  &  \textbf{0.76}  &  \textbf{0.96}  &  \textbf{83.33} \\
004\_sugar\_box  &  1890.38  &  0.60  &  5.75  &  1.71  &  1.77  &  66.15  &  \textbf{717.21} &  \textbf{0.48}  &  \textbf{2.19}  &  \textbf{1.01}  &  \textbf{1.26}  &  \textbf{100.00} \\
005\_tomato\_soup\_can  &  1874.24  &  0.64  &  6.60  &  1.85  &  \textbf{1.62}  &  64.84  &  \textbf{537.16} &  \textbf{0.34}  &  \textbf{1.71}  &  \textbf{1.44}  &  1.66  &  \textbf{83.33} \\
006\_mustard\_bottle  &  1909.55  &  0.94  &  5.04  &  1.45  &  \textbf{1.47}  &  32.81  &  \textbf{652.70} &  \textbf{0.65}  &  \textbf{2.26}  &  \textbf{1.34}  &  1.60  &  \textbf{83.33} \\
007\_tuna\_fish\_can  &  1730.41  &  0.66  &  6.21  &  1.70  &  1.54  &  \textbf{66.67}  &  \textbf{499.62} &  \textbf{0.22}  &  \textbf{1.33}  &  \textbf{0.50}  &  \textbf{0.60}  &  50.00 \\
008\_pudding\_box  &  1620.01  &  0.88  &  6.69  &  1.86  &  1.75  &  83.33  &  \textbf{530.88} &  \textbf{0.64}  &  \textbf{2.30}  &  \textbf{0.92}  &  \textbf{1.12}  &  83.33 \\
009\_gelatin\_box  &  1601.29  &  0.99  &  2.46  &  \textbf{0.88}  &  \textbf{1.10}  &  0.00  &  \textbf{497.16} &  \textbf{0.40}  &  \textbf{1.51}  &  1.04  &  1.20  &  \textbf{50.00} \\
010\_potted\_meat\_can  &  1402.83  &  0.71  &  7.24  &  1.67  &  \textbf{1.56}  &  52.60  &  \textbf{610.07} &  \textbf{0.46}  &  \textbf{2.33}  &  \textbf{1.56}  &  1.80  &  \textbf{66.67} \\
011\_banana  &  1899.47  &  0.75  &  5.67  &  2.06  &  1.91  &  66.67  &  \textbf{497.27} &  \textbf{0.36}  &  \textbf{2.62}  &  \textbf{1.39}  &  \textbf{1.56}  &  66.67 \\
019\_pitcher\_base  &  2123.26  &  0.81  &  5.96  &  1.94  &  2.07  &  46.21  &  \textbf{725.94} &  \textbf{0.58}  &  \textbf{1.52}  &  \textbf{0.90}  &  \textbf{1.17}  &  \textbf{66.67} \\
021\_bleach\_cleanser  &  1512.40  &  \textbf{0.63}  &  6.67  &  \textbf{1.23}  &  \textbf{1.38}  &  16.67  &  \textbf{824.98} &  0.88  &  \textbf{2.50}  &  1.62  &  1.96  &  \textbf{83.33} \\
024\_bowl  &  2400.90  &  \textbf{0.81}  &  9.03  &  1.94  &  1.89  &  \textbf{90.10}  &  \textbf{498.04} &  0.86  &  \textbf{5.14}  &  \textbf{0.93}  &  \textbf{1.23}  &  83.33 \\
025\_mug  &  2300.23  &  0.85  &  5.64  &  1.45  &  1.57  &  50.00  &  \textbf{529.74} &  \textbf{0.42}  &  \textbf{2.27}  &  \textbf{0.52}  &  \textbf{0.71}  &  50.00 \\
035\_power\_drill  &  1097.89  &  1.80  &  \textbf{1.91}  &  \textbf{1.26}  &  \textbf{1.53}  &  0.00  &  \textbf{565.60} &  \textbf{0.53}  &  2.06  &  1.99  &  2.22  &  \textbf{83.33} \\
036\_wood\_block  &  1421.61  &  \textbf{0.60}  &  3.62  &  1.36  &  1.58  &  16.67  &  \textbf{503.09} &  0.79  &  \textbf{2.56}  &  \textbf{1.26}  &  \textbf{1.57}  &  \textbf{100.00} \\
037\_scissors  &  2398.00  &  3.43  &  22.70  &  2.34  &  1.83  &  25.00  &  \textbf{489.47} &  \textbf{0.23}  &  \textbf{1.02}  &  \textbf{0.30}  &  \textbf{0.39}  &  \textbf{33.33} \\
040\_large\_marker  &  1016.87  &  0.75  &  22.33  &  1.64  &  1.65  &  \textbf{16.93}  &  \textbf{459.93} &  \textbf{0.07}  &  \textbf{0.48}  &  \textbf{0.15}  &  \textbf{0.16}  &  16.67 \\
052\_extra\_large\_clamp  &  2179.48  &  0.92  &  7.55  &  2.00  &  1.58  &  \textbf{67.77}  &  \textbf{463.35} &  \textbf{0.20}  &  \textbf{1.36}  &  \textbf{0.27}  &  \textbf{0.36}  &  33.33 \\
061\_foam\_brick  &  1694.44  &  0.85  &  5.78  &  2.09  &  1.70  &  73.08  &  \textbf{485.18} &  \textbf{0.47}  &  \textbf{2.61}  &  \textbf{1.44}  &  \textbf{1.68}  &  \textbf{83.33} \\
\hline
Total(Avg.)  &  1823.51  &  0.84  &  7.60  &  1.66  &  1.63  &  45.91  &  \textbf{572.29}  &  \textbf{0.48}  &  \textbf{2.01}  &  \textbf{1.04}  &  \textbf{1.25}  &  \textbf{64.58} \\
\hline\hline
\end{tabular}
\caption{\textbf{Per-object quantitative comparison on DexYCB.} 
Each row corresponds to a distinct YCB object, so the table reveals how performance varies with object geometry (e.g., cans, thin bowls, elongated tools). 
Across most objects, our method is about $3{\sim}4\times$ faster to optimize than ManipTrans and consistently achieves lower position errors ($E_t, E_j, E_{ft}$) and comparable or better rotation error $E_r$, resulting in higher success rates even for challenging shapes. 
(Note: $E_t, E_j, E_{ft}$ are in cm and $E_r$ is in degrees.)}

\label{tab:quant_per_object}
\end{table*}

\subsection{Results}
\label{subsec:results}

We evaluated our method against the MANIPTRANS baseline on our curated DexYCB test set. For the policy-based MANIPTRANS, we followed their protocol by performing 100 rollouts per sequence and computing error metrics \emph{only for the successful rollouts}. In contrast, our deterministic framework runs one optimization per sequence.

As shown in Table~\ref{tab:quant_per_subject} and Table~\ref{tab:quant_per_object}, our method significantly outperforms the baseline across nearly all evaluation metrics. Notably, our approach achieves this superior accuracy while requiring considerably less computation time. These results indicate that our optimization framework can efficiently discover physically-executable hand poses that not only stably grasp and lift objects but also closely track the object trajectories in the input images. 

Our method's strong performance is largely driven by its robustness to data noise. MANIPTRANS, which was trained on highly accurate motion capture data, appears to struggle with the noisier pose estimates inherent in the DexYCB dataset, leading to performance degradation. Our optimization-based framework demonstrates greater robustness to this noise.

While our method is more robust, we still observe some performance variance across subjects (Table~\ref{tab:quant_per_subject}). We hypothesize this stems from the \emph{morphological gap} between the subjects' real hands and the fixed-scale Arti-MANO simulation model. This discrepancy, particularly in finger length, can make specific grasps challenging, though our method still successfully handles complex motion sequences where the baseline often fails. 

%

\subsection{Qualitative Analysis of Physics Reconstruction}
\label{subsec:qualitative}

A key benefit of our framework is the reconstruction of hidden physical phenomena, such as contact positions and forces, from vision-only data. The original vision-aligned trajectories, while visually accurate, are often physically inconsistent. They exhibit missed contacts, interpenetration, or unstable grasps that cause objects to drift or fall when replayed in simulation (see Fig.~\ref{fig:objective}).

\begin{figure*}[t]
    \centering
    \includegraphics[width=\linewidth]{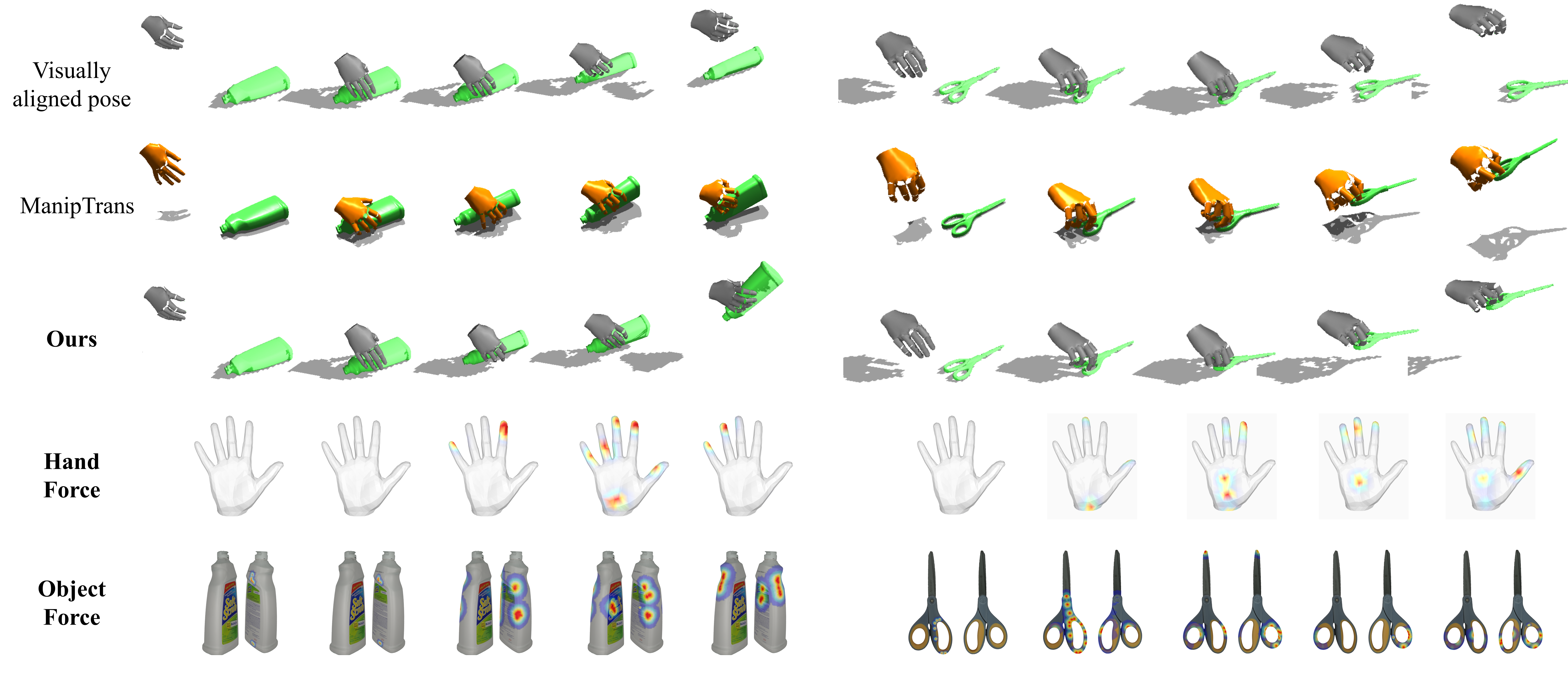}
    \caption{\textbf{Qualitative comparison.}
Top row: original vision-aligned hand--object poses from DexYCB~\cite{Chao2021DexYCB}, which often contain gaps or interpenetration.
Second row: retargeted trajectories produced by ManipTrans, which can still exhibit unstable or physically implausible grasps.
Third row: trajectories refined by our method, yielding more natural motion and physically plausible, stable grasps that remain consistent with the image evidence.
The bottom two rows visualize per-vertex contact force magnitudes on the hand and object obtained from MuJoCo, showing that our physics-based refinement exposes dense force information unavailable in the original dataset or in purely kinematic baselines.}

    \label{fig:qualitative}
\end{figure*}

Our optimization explicitly resolves these issues by restoring two key properties:
\begin{itemize}
\item Geometric Plausibility: The refined hand trajectory is adjusted to form stable grasps at plausible fingertip and palm locations. This eliminates the penetrations and ``floating" grasps common in the original vision-only hand and object trajectories.
\item Dynamic Plausibility: We observe this in both ``grasp-and-lift" and ``press-and-stabilize" sequences. Where unrefined replays show negligible influence or unrealistic impulses, our method applies consistent, stable pressure. The recovered contact forces are sufficient to support the object against gravity, and the resulting motion is both stable and faithful to the visually inferred trajectory.
\end{itemize}
Overall, our method successfully turns vision-only poses into replayable, physically-grounded interaction data, restoring both geometric and dynamic plausibility while generating the underlying contact forces that explain the motion.
See Fig.~\ref{fig:qualitative} for visualizations of the reconstructed hand-object trajectories, contact positions and forces, which confirm that our method recovers realistic and physically stable grasp-and-lift patterns.




\begin{table}[t]
\centering

\small
\begin{tabular}{@{~}l@{~~}c@{~~}c@{~~}c@{~~}c@{~~}c@{~~}r@{~}}
\toprule
& Time (s) & $E_t$ & $E_r$ & $E_j$ & $E_{ft}$ & SR (\%) \\
\midrule
\textbf{Ours} & 
\textbf{529.80} & \textbf{0.43} & \textbf{1.99} & \textbf{0.47} & \textbf{0.59} & \textbf{100.00} \\

\textbf{Ours w/o $\cL_\text{obj-vel}$} & 
688.07 & 0.53 & 2.08 & 0.88 & 1.02 & 96.78 \\

\textbf{Ours w/o $\cL_\text{hand-joint}$} & 
619.51 & 0.73 & 2.18 & 1.40 & 1.94 & 93.33 \\
\bottomrule
\end{tabular}
\caption{\textbf{Ablation on our objective losses.} 
\emph{Ours} denotes the full optimization pipeline, and each variant zeroes out the corresponding loss weight to isolate its effect on the final optimization performance. Refer to the text for the experiment setup.}

\label{tab:ablation}
\end{table}

\medskip
\subsection{Ablation}
\label{subsec:ablation}
\medskip

We performed ablation studies on our two key objective terms: the Object Velocity Loss ($\cL_\text{obj-vel}$) and the Hand 3D Joint Position Loss ($\cL_\text{hand-joint}$). To create a controlled comparison, we selected 30 sequences from Subject 01 on which our full method (Ours) achieved a 100\% success rate. We then ran each ablation variant on this specific subset to isolate the impact of removing each component. The results are summarized in Tab.~\ref{tab:ablation}

\paragraph{Object Velocity Loss ($\cL_\text{obj-vel}$).} Removing this loss forces the optimizer to rely only on per-frame static object poses. As shown in Tab.~\ref{tab:ablation}, this degrades all error metrics and the success rate, while also increasing optimization time. While the optimizer can still match static poses, it ignores \emph{the trajectory-level dynamics} (consistency of linear and angular velocities) encoded by consecutive frames. Qualitatively, this leads to dynamically inconsistent motions: objects may exhibit small jitter, delayed responses, or overly ``stiff" motion. By explicitly matching velocities, $\cL_\text{obj-vel}$ provides a strong temporal prior on how the object should move, not just \emph{where} it should be, leading to more reliable reconstructions.

\paragraph{Hand 3D Joint Position Loss $\cL_\text{hand-joint}$.}
This ablation tests the hypothesis that the optimizer should focus only on the object's motion, not the hand joints. However, Tab.~\ref{tab:ablation} shows this approach is highly detrimental, significantly degrading all metrics, including object translation error ($E_t$), and reducing the success rate by over 6\%. This indicates that the image-aligned hand keypoints provide a crucial visual prior on how the hand should approach and support the object. Without this prior, the optimizer can find unnatural hand configurations (e.g., twisted wrists) that still move the object, leading to larger errors and less stable interactions. $\cL_\text{hand-joint}$ is therefore essential for anchoring the search to grasps that are both visually consistent and physically plausible.

\section{Conclusion and Future Work}
\label{sec:conclusion}

We presented a physics-constrained trajectory refinement framework that successfully transforms vision-only hand-object demonstrations into physically-executable interactions. Our core contribution is a tractable, black-box optimization scheme. By parameterizing the hand's motion with a spline-based temporal representation and optimizing only the hand's control inputs with CMA-ES, we allow the object's motion to emerge directly from the simulated physical interaction with hand and fingers. This approach resolves the ``vision-physics mismatch," eliminating penetrations and stabilizing grasps. Empirically, our method achieves lower trajectory errors and faster convergence than the state-of-the-art MANIPTRANS, all while generating explicit contact forces and positions as a free by-product. 

Despite these benefits, our approach has limitations that open promising directions for future work. The refinement is an offline, per-sequence procedure that currently assumes accurate, rigid object geometry and a reasonably well-tracked object pose; large reconstruction errors or deformable objects remain challenging. 
A natural next step is to use our pipeline to build large-scale, contact-annotated datasets from diverse vision-only recordings. This refined, physically-grounded data can then serve as a valuable new source of supervision for training contact-rich robot manipulation policies, such as grasping, in-hand manipulation, and tool use without relying on force-torque sensors. Extending the framework to bimanual settings, different robot embodiments, and more complex contact graphs is also an exciting avenue for future research.

{
    \small
    \bibliographystyle{ieeetr}
    \bibliography{main}
}

\clearpage
\setcounter{page}{1}
\appendix
\maketitlesupplementary

\begin{figure*}[t!]
    \centering
    \includegraphics[width=\linewidth]{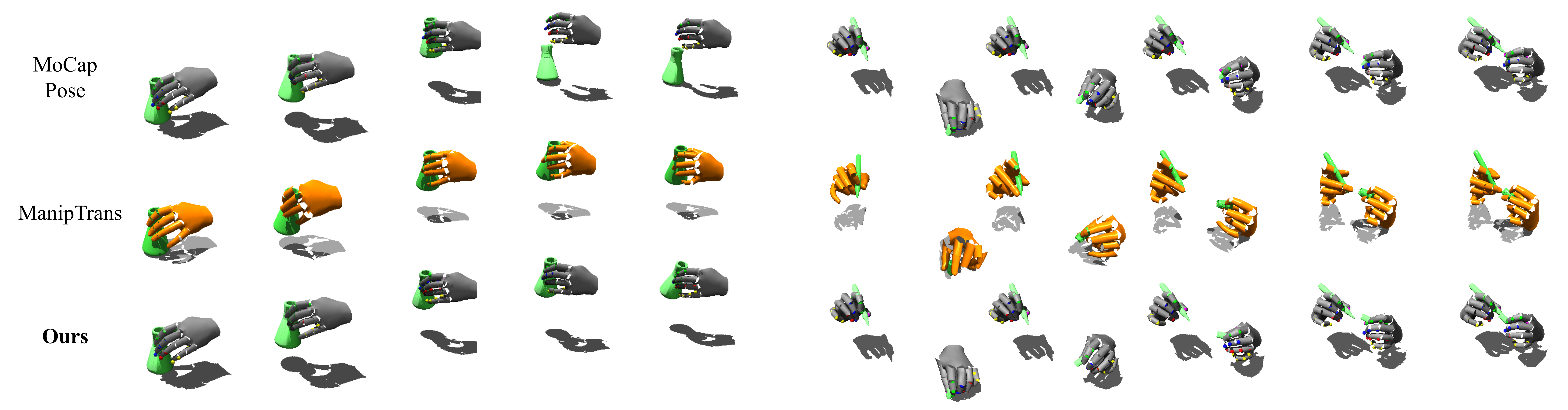}
    \caption{\textbf{Qualitative comparison on OakInk-V2~\cite{OAKINK2} dataset.}
Top row: original ground truth hand trajectory from OakInk-V2.
Second row: retargeted trajectories produced by ManipTrans.
Third row: trajectories refined by our method.}

    \label{fig:qualitative_oakink}
\end{figure*}

\section{Results on Other Hand–Object Datasets}

To demonstrate the robustness of our method, we further validated it on the OakInk-V2~\cite{OAKINK2} dataset, which was primarily used in ManipTrans~\cite{Li2025ManipTrans}. For bimanual manipulation tasks, we employed independent CMA-ES instances for each hand to simplify the implementation, sampling the poses separately. Following the protocol in ManipTrans, we utilized the OakInk-V2 dataset at 60Hz. Consequently, we adjusted the keyframe interval from 5 to 10 to maintain temporal consistency.

The qualitative results, as shown in Fig.~\ref{fig:qualitative_oakink}, indicate that our method successfully optimizes motion capture demonstrations. Compared to ManipTrans, our approach yields significantly smoother trajectories. By generating intermediate control parameters via cubic spline interpolation, we effectively mitigate high-frequency artifacts, ensuring temporal consistency for both single-hand and bimanual tasks (best viewed in the supplementary video). However, it should be noted that the optimization time scales linearly with the number of frames.

\section{Reconstructed Physical Hand--Object Interactions}
\medskip

\subsection{Physical Interactions}
\medskip

Original vision-aligned trajectories, optimized only for image consistency, often contain physical artifacts such as hovering fingertips or deep interpenetration, leading to simulation failure (Fig.~\ref{fig:objective}). Our physics-in-the-loop refinement converts these into replayable sequences with geometrically valid contacts. The refined rollouts closely track the original visual trajectory while ensuring physically plausible grasp and manipulation behavior.

\begin{figure*}[t]
    \centering
    \includegraphics[width=\linewidth]{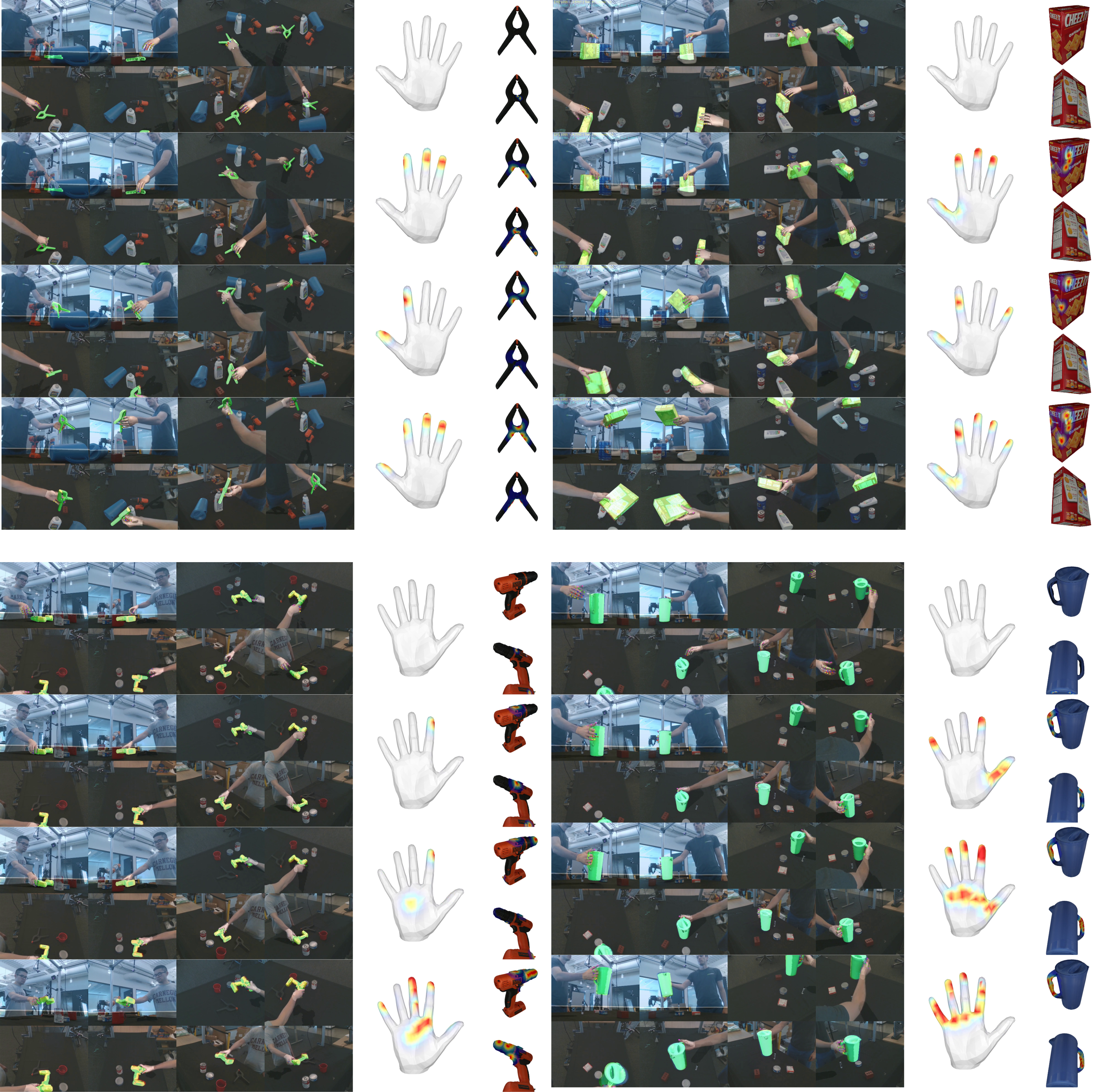}
    \caption{\textbf{Additional examples across diverse scenes.}
    We show multiple DexYCB sequences containing different objects and grasp types. 
    For each sequence, our refinement produces physically valid hand--object interactions and yields diverse, object-dependent contact locations and force patterns.}
    \label{fig:diverse_scene}
\end{figure*}

Fig.~\ref{fig:banana_bowl} shows this improvement in problematic scenes. The noticeable hovering in the banana example and severe interpenetration in the bowl example are resolved into stable, physically valid grasps. These results show how optimization bridges visual–physical inconsistencies to reconstruct realistic interactions. Further examples in Fig.~\ref{fig:diverse_scene} confirm the pipeline's robustness across diverse object geometries and interaction types.

\medskip
\subsection{Contact Forces}
\medskip

\begin{figure*}[t]
    \centering
    \includegraphics[width=\linewidth]{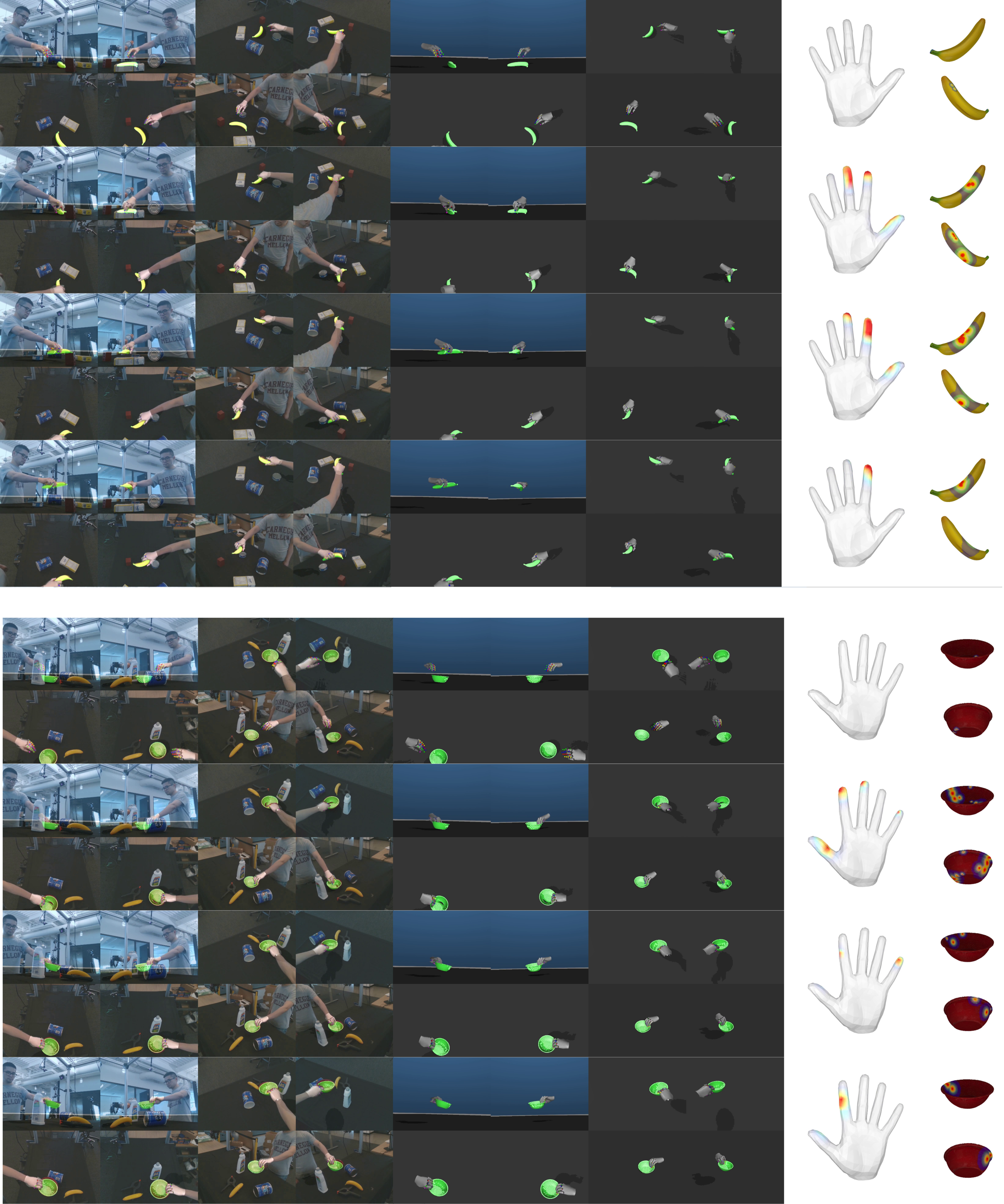}
    \caption{\textbf{Reconstructed physical hand--object interactions.}
    Left: ground-truth RGB frames overlaid with the optimized simulated hand and object meshes, visualized on both the hand and the object.
    Middle: replay of the same optimized hand trajectory in the physics simulator using only the interactable object.
    Right: contact information visualized (contact locations and force magnitudes).
    The top four rows show a banana scene and the bottom four rows a bowl scene. In both cases, refinement eliminates the original fingertip--object gaps and severe interpenetration, yielding physically plausible trajectories and stable interactions.}
    \label{fig:banana_bowl}
\end{figure*}

With the trajectories refined into realistic motions, we can directly query the simulator for dense contact information. For every frame, we extract time-aligned contact positions, normals, and 6-DoF wrenches on both the hand and the object, as visualized in the right column of Fig.~\ref{fig:banana_bowl}. This transforms each refined sequence into a richly annotated record of physical interaction rather than a purely kinematic track. These contact forces serve as valuable supervision for downstream tasks, including learning contact-aware policies, analyzing grasp stability, and force estimation.

\begin{figure*}[t]
  \centering
  \includegraphics[width=\textwidth]{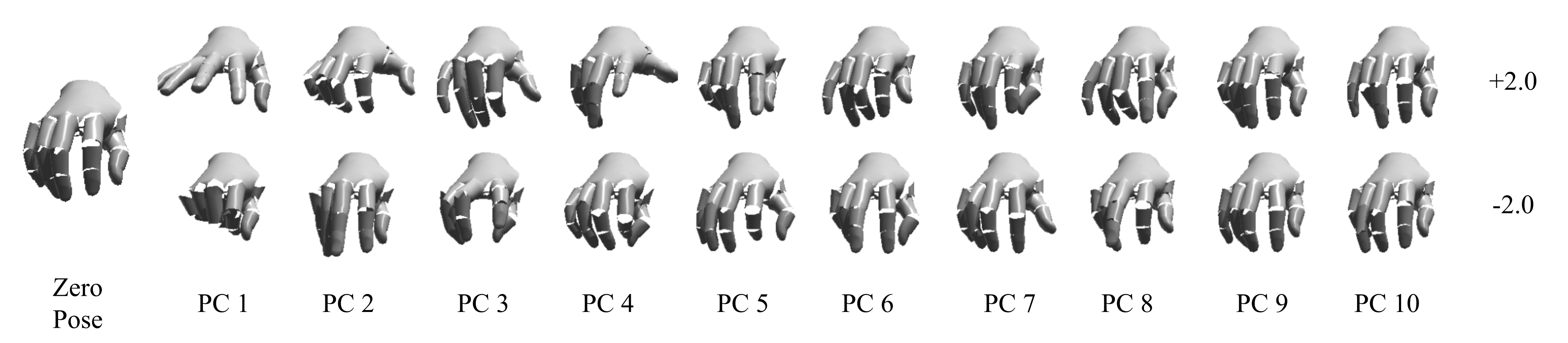}
  \caption{The PCA pose space of Arti-MANO~\cite{Christen2022DGrasp} hand model. The left-most image depicts the zero pose, while the remaining columns illustrate the effect of the first ten principal components (PCs). The effect of each PC is visualized by adding $\pm 2$ standard deviation (std) to the mean pose, as indicated.}
  \label{fig:pca_figure}
\end{figure*}

\section{Implementation Details}
\medskip

\subsection{Framework Loss Weights}
\medskip
We optimize a weighted sum of several losses: an object pose loss $\mathcal{L}_{\text{obj}}$ on the SE(3) pose of the manipulated object, a hand pose loss $\mathcal{L}_{\text{hand}}$ on 3D MANO keypoints (with fingertips up-weighted), an object velocity loss $\mathcal{L}_{\text{obj-vel}}$ matching the 6-DoF velocity of the simulated object to finite-difference velocities from the visually aligned object trajectory, a contact distance loss $\mathcal{L}_{\text{contact}}$ between fingertips and the inferred object contact points, and two smoothness regularizers $\mathcal{L}_{\text{cons}}$ (frame-to-frame control consistency) and $\mathcal{L}_{\text{hand-vel}}$ (hand body velocities).

All distance- and velocity-based terms are expressed in metric units (meters or m/s), while the rotational part of $\mathcal{L}_{\text{obj}}$ is measured in radians and mixed with translation using a small rotation-to-translation ratio so that centimeter-level pose errors dominate over small orientation discrepancies. In our main 3D keypoint setting, the largest weights are assigned to $\mathcal{L}_{\text{obj}}$ and $\mathcal{L}_{\text{hand}}$, with $\mathcal{L}_{\text{obj-vel}}$ on a comparable scale to suppress object jitter; the contact term $\mathcal{L}_{\text{contact}}$ is given a moderate weight (roughly a few times smaller than the pose terms), and the smoothness terms $\mathcal{L}_{\text{cons}}$ and $\mathcal{L}_{\text{hand-vel}}$ use weights that are two or three orders of magnitude smaller, serving only as soft regularizers.

\medskip
\subsection{MuJoCo Implementations}
\medskip

All refinements are run with a fixed timestep of $0.00416$\,s (240\,Hz), using the \texttt{implicitfast} integrator and a Newton solver with 10 iterations and 10 \texttt{noslip\_iterations}. Collisions are enabled only between the hand and the object (and between the object and the table). We use a single set of soft-contact parameters for all colliding geoms, with \texttt{solimp} set to $(0.9, 0.99, 0.001)$ and \texttt{solref} set to $(0.003, 1)$, which provides stable sticking contacts while avoiding excessive penetration.

For each YCB object, we take the mass directly from the DexYCB specification (e.g., $0.066$\,kg for the banana) and compute the diagonal inertia from the collision mesh and this mass, so that the simulated dynamics are consistent with the dataset. The modified Arti-MANO wrist is rigidly welded to a kinematic control body that we directly drive along the optimized vision-aligned trajectory, using \texttt{solimp} $(0.8, 0.95, 0.01)$ and \texttt{solref} $(0.01, 1)$. This non-physical control body allows us to reposition the wrist exactly as desired (effectively “teleporting” it) while the rest of the hand remains fully simulated and interacts physically with the object. All finger joints are actuated by position servos with uniform gains $k_p = 50$, $k_v = 3$, and joint control ranges of roughly $[-1, 1]$, so that the modified Arti-MANO hand can faithfully track arbitrary PCA poses; these settings are kept fixed across all experiments.

Fig.~\ref{fig:pca_figure} illustrates the variations in the Arti-MANO hand model corresponding to each principal component (PC). As shown in the figure, the initial components induce the most significant geometric changes, while the influence of subsequent components progressively diminishes. This observation suggests that the first 10 PCs are sufficient to represent the majority of plausible hand poses.

\end{document}